\documentclass[letterpaper]{article} 
\usepackage[]{aaai23}  
\usepackage{times}  
\usepackage{helvet}  
\usepackage{courier}  
\usepackage[hyphens]{url}  
\usepackage{graphicx} 
\usepackage{natbib}  
\usepackage{caption} 
\usepackage{amsmath}
\usepackage{booktabs} 
\usepackage{subcaption}
\usepackage{multirow}
\usepackage{multicol}
\usepackage{bm}
\usepackage[ruled,noend]{algorithm2e}
\usepackage{xcolor}
\usepackage{array}
\usepackage{varwidth}
\usepackage{longtable}
\DontPrintSemicolon

\usepackage{soul}

\title{Towards Automatic Evaluation of Task-Oriented Dialogue Flows}
\author {
    Mehrnoosh Mirtaheri,\textsuperscript{\rm 1, 2, \footnote{Work was conducted during the summer internship in 2022.}}
    Nikhil Varghese, \textsuperscript{\rm 2}
    Chandra Khatri, \textsuperscript{\rm 2}
    Amol Kelkar \textsuperscript{\rm 2}
}
\affiliations {
    \textsuperscript{\rm 1} USC Information Sciences Institute\\
    \textsuperscript{\rm 2} Got It AI\\
    mehrnoom@usc.edu, nikhil@bot-it.ai, chandra@bot-it.ai, amol@bot-it.ai \\

}

\begin{document}

\maketitle

\begin{abstract}


Task-oriented dialogue systems rely on predefined conversation schemes (dialogue flows) often represented as directed acyclic graphs. These flows can be manually designed or automatically generated from previously recorded conversations. Due to variations in domain expertise or reliance on different sets of prior conversations, these dialogue flows can manifest in significantly different graph structures. Despite their importance, there is no standard method for evaluating the quality of dialogue flows. We introduce FuDGE (Fuzzy Dialogue-Graph Edit Distance), a novel metric that evaluates dialogue flows by assessing their structural complexity and representational coverage of the conversation data. FuDGE measures how well individual conversations align with a flow and, consequently, how well a set of conversations is represented by the flow overall. Through extensive experiments on manually configured flows and flows generated by automated techniques, we demonstrate the effectiveness of FuDGE and its evaluation framework. By standardizing and optimizing dialogue flows, FuDGE enables conversational designers and automated techniques to achieve higher levels of efficiency and automation.

\end{abstract}

\section{Introduction}
\begin{figure}[t]
    \centering
    \includegraphics[width=0.9\columnwidth]{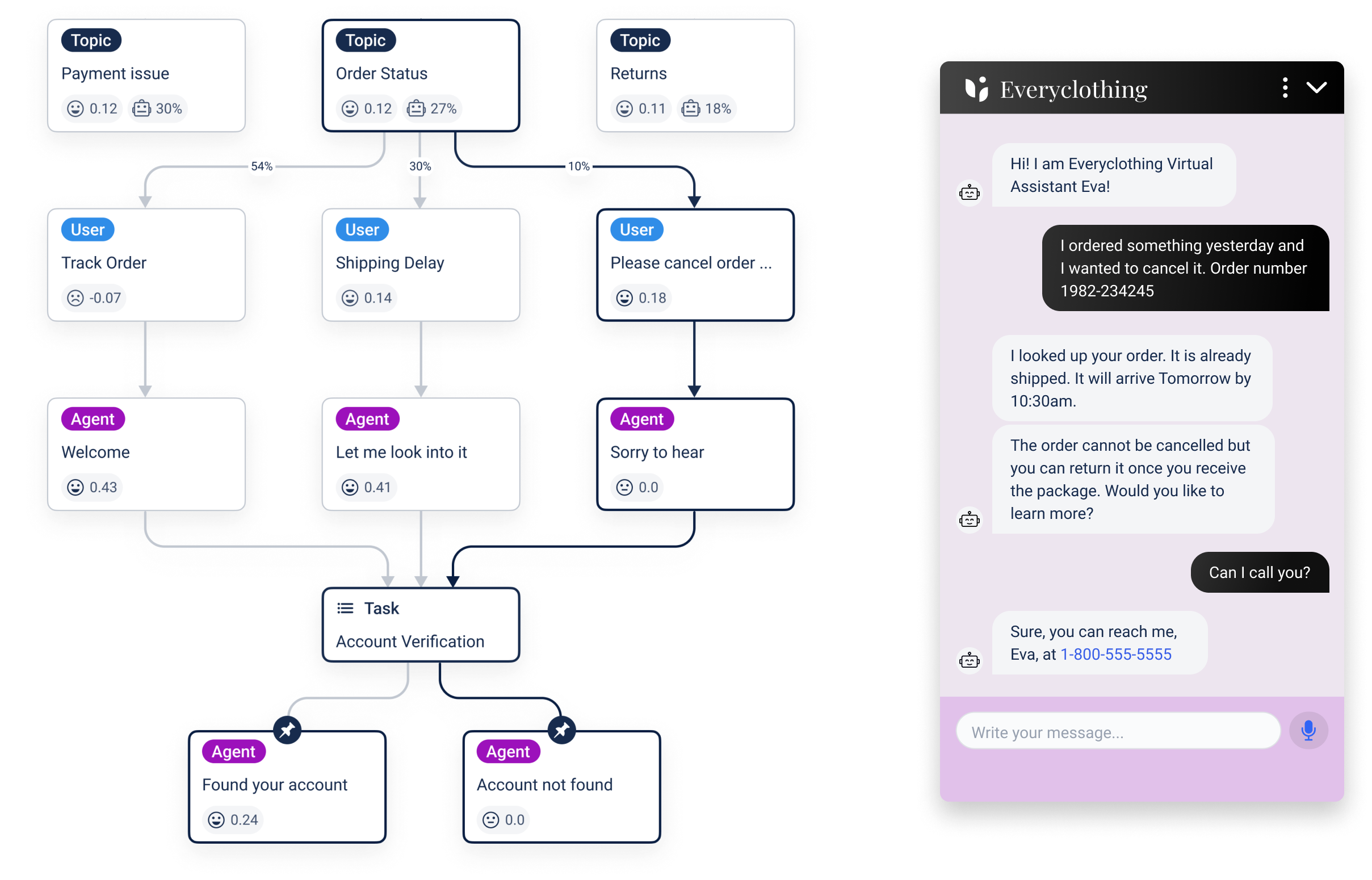}
    \caption{An illustration of a dialogue flow (left) that is used to configure a task-oriented dialogue agent for a fictitious company named "Everyclothing." The chat widget (right) illustrates a customer asking about canceling their order. }
    \label{fig:gotit}
\end{figure}

One of the most promising applications of Conversational AI lies in Customer Service Automation, where task-oriented dialogue systems aim to address customer concerns effectively without human intervention. These systems often rely on dialogue flows—structured conversation schemes—to retrieve appropriate information from knowledge bases or back-end systems. Over the years, frameworks like Dialogflow CX\footnote{https://cloud.google.com/dialogflow/cx/docs/basics}, Rasa\footnote{https://rasa.com}, and Amazon Lex\footnote{https://aws.amazon.com/lex/} have facilitated the creation of task-oriented dialogue systems by leveraging flows comprising user intents, agent actions, and other metadata. As shown in Figure \ref{fig:gotit}, dialogue flows define possible paths a customer and agent can take during a conversation, enabling structured automation.

Dialogue flows are typically handcrafted by domain experts through an iterative process involving historical data analysis. This iterative process depends on the designer’s expertise, the quality of the data, and the time invested, resulting in dialogue flows or directed acyclic graph (DAG) that vary significantly in flow attributes such as the number of user intents, intent sequences, and graph paths. 

Alternatively dialogue flows can be can be generated using automated flow discovery algorithms, which leverage historical dialogue data. These algorithms rely on hyperparameters that significantly influence the resulting flow structure. However, evaluation of flows with respect to historical dialogues has been a relatively unexplored area of NLP. Previous work has predominantly focused on improving the quality of task-oriented dialog agents through better intent discovery \cite{zhang2022new, shi-etal-2018-auto, perkins-yang-2019-dialog} which is often categorized as a sub-problem in the dialogue flow discovery process. The domain of flow discovery from historical human-human dialogues has had relatively sparse research despite being crucial for building task-oriented dialogue agents. There is a need to automatically evaluate the quality of the dialogue flows with respect to the historical dialogue transcripts used to generate them to deliver a consistent baseline, better versioning and tracking progress of flows, and corresponding automation with time.

To that end, we introduce a novel evaluation framework to assess the flow discovery algorithm or to guide the human conversation designer while building the flows. Our evaluation framework enables comparison between the flows generated from the same corpus by assigning a score to a dialogue flow. A good flow should cover the important and representative conversations from the corpus while simultaneously being as compressed as possible. Note that there is a trade-off between the degree of compression and the amount of information being lost. A graph containing all conversations potentially covers the entire corpus, but it is not beneficial for designing a dialogue system because it will strictly imitate historical conversations and will not generalize well to unseen dialogues. On the other hand, a simple graph \texttt{start -> Hello -> Goodbye} is still a dialogue flow, but it loses most of the information about the corpus. Our evaluation framework takes both of these aspects into consideration. As the core of our evaluation framework, we propose FuDGE, \textbf{Fu}zzy \textbf{D}ialogue-\textbf{G}raph \textbf{E}dit distance, an algorithmic way of computing the distance between a specific conversation and a given dialogue flow paths. FuDGE allows us to measure how well the dialogues are represented and covered by a dialogue flow. We combine the FuDGE score with the complexity of the DAG representing the flow and produce \textbf{Flow}-\textbf{F}1 (\textit{FF1}) score that captures the trade-off between the compression and complexity. 

Overall, our contributions are as follows:
\begin{enumerate}
    \item We introduce a novel  dialogue flow evaluation framework that provides a metric to quantify the goodness of a dialogue flow generated from a dialogue corpora, taking into consideration the inherent trade-off between flow complexity and information contained in the flows
    \item We propose FuDGE distance, an efficient edit-distance metric between a dialogue flow and a conversation that can independently be used in any dialogue evaluation task, including zero-shot dialogue generation from predefined schemes. We show that FuDGE can effectively separate within-task versus out-of-task conversations for a given set of dialogues.
    \item{We demonstrate that \textit{FF1} score can help hyperparameter tuning, ranking, and pruning the dialogue flows to identify the most optimal flows for a given dialogue corpus.}
\end{enumerate}

\section{Preliminaries}
In this section, we begin by defining the important terms frequently used throughout the paper involving the dialogue flows and describe the core idea behind the automatic dialogue flow discovery methods. Then, we explain the Levenshtein distance \cite{levenshtein1966binary} and the algorithm to compute it as a foundation for the FuDGE algorithm.

\subsection{Automatic Flow Discovery}
Automatic flow discovery aims to obtain a dialogue flow from a dialogue corpus. 
A dialogue corpus consists of a set of $N$ dialogues. Each dialogue includes a sequence of agent and user utterances that occur turn by turn. A dialogue flow is a DAG representing the corpus in the form of nodes, where each node is a user intention/request or an agent response. 

A naive way of obtaining a graph from a dialogue corpus is assigning a node to every utterance and connecting the consecutive utterance nodes. This results in a massive graph with the same number of nodes as the number of utterances. This is impractical for initializing a dialogue agent or giving to a human for manual configuration because it would capture all the noise inherent in human-human conversations. Also, the graph would be extremely large for practical, real-world dialogue datasets, making it very hard for manual configuration or automation. To achieve more condensed representations, automatic flow discovery methods broadly follow two steps:
\begin{enumerate}
    \item Identify the type of agent responses and user requests and assign an intent label to each user/agent utterance. An intent label \texttt{"book hotel"} might include user utterances like \texttt{"I want to book a hotel"}, \texttt{"Need a room in the Marriott for next month"}, and \texttt{"Need accommodation this weekend"}. This could be achieved by training a classifier on manually annotated user and agent utterances or by performing density-based clustering like DBSCAN \cite{ester1996density} using pre-trained encoders like BERT \cite{devlin2018bert}. Various semi-supervised methods \cite{forman2015clustering, lin2020discovering, zhang2021discovering} also focus on the intent discovery task, which is to find an utterance intent in the presence of some existing intents.
    \item Once the dialogue utterances are replaced by intent labels, automatic flow discovery methods employ various AI techniques to convert the conversation paths into a more condensed graph. Some methods may facilitate flow ranking strategies that discover the most important paths and pruning strategies that remove the less important nodes or edges from the graph. 
\end{enumerate}

\noindent In this work, we use two proprietary automatic flow discovery methods, each employing different strategies to build the dialogue flows. The final expected outcome is a dialogue flow similar to what is depicted in Figure \ref{fig:gotit}. For both algorithms, we can use annotated data (agent and user utterance intent labels) or a clustering method to assign labels to the utterances. In our experiments, we generate flows with and without intent labels to demonstrate the efficacy of our work.

\subsection{Levenshtein Distance}
The Levenshtein distance (aka edit distance) is the backbone of numerous search algorithms. It is defined as the minimum number of \textit{deletion}, \textit{insertion} or \textit{substitution} operations needed to convert a string $a$ to string $b$, where $a=a_1a_2a_3\dots a_n$ and $b=b_1b_2b_3\dots b_m$ are sequences of single characters. Given the sequence $a$ and $b$, we define $d_{r,s}$ to be the minimum number of operations used to convert sub-strings $a_{1:r} = a_1a_2\dots a_r$ to $b_{1:s} = b_1b_2 \dots b_s$. $d_{rs}$ can be recursively computed by considering three following possibilities:
\begin{enumerate}
    \item delete $a_r$, then match $a_{1:r-1}$ with $b_{1:s}$. 
    \item insert $b_s$ at the end of $a_{1:r}$, then match $a_1a_2\dots a_r b_s$ with $b_1b_2 \dots b_{s-1} b_s$ or equivalently $a_{1:r}$ and $b_{1:s-1}$.
    \item substitute $a_r$ with $b_s$, then match $a_{1:r-1}$ and $b_{1:s-1}$.
\end{enumerate}

The minimum number of operations $d_{r,s}$ is the minimum cost of the three previous steps:

\begin{equation}
\label{eq:Edit_Distance}
    d_{r,s}= min \left\{
  \begin{array}{@{}ll@{}}
    d_{r-1,s} + c_{del}(a_r), & \text{case. 1} \\
    d_{r,s-1} + c_{ins}(b_s), & \text{case. 2} \\
    d_{r-1,s-1} + c_{sub}(a_r, b_s), & \text{case. 3}
  \end{array}\right.
\end{equation}

Where $c_{del}$,  $c_{ins}$, $c_{sub}$ are the deletion, insertion and substitution cost respectively. $ c_{sub}(a_r, b_s)$ is 0 if $a_r = b_s$ and 1 otherwise. The initial cases, $d_{0,s}$, the cost of converting an empty string to a full string, and $d_{r, 0}$, the cost of converting a full string to an empty string, are defined as:
\begin{equation}
     d_{0,s} = \sum_{t=0}^s c_{ins}(b_t), d_{r,0} = \sum_{s=0}^r c_{del}(a_t)
\end{equation}

 The final edit distance between $a$ and $b$ is $d_{m, n}$, and it can be calculated efficiently using dynamic programming with complexity $\mathcal{O}(mn)$e. Furthermore, tracking the steps at each iteration gives us the \textit{alignment} between two strings, where each character of the first string is either matched with another character or a gap in the other string.

\section{Problem Definition}
We represent a dialogue flow with a DAG $G = (V, E)$, and its root node $G_r$. Each path in $G$ starting from the root corresponds to a possible user-agent interaction scenario in a dialogue system. Each node in $G$ is associated with an intent bucket $B^i \in B = \{B^1, B^2, \dots, B^M\}$. A bucket $B^i = (actor, utterances)$ consists of a collection of utterances, grouped together if they all convey the same semantic corresponding to a user intent ($actor = user$) or an agent action ($actor = agent$). Throughout the paper, we use the term intent for both users and agents. For simplicity, we also assume that the root node $G_r$ is the start of all the conversations and is associated with a dummy bucket. A flow path $P = G_rP_1P_2...P_n$ is a path in graph $G$, where $P_i \in V$ and $(P_i, P_{i+1})$ is an edge in $E$. 

To generate a dialogue flow, human experts or dialogue flow discovery methods leverage a dialogue corpus $C = \{C^1, C^2, \dots, C^N \}$ consisting of $N$ recorded conversations between users and agents in a service center. The dialogue flows obtained from a dialogue corpus can vary significantly, depending on the biases of human experts and machine algorithms. Therefore it is essential to devise an automatic way of comparing various dialogue flows obtained from a corpus. 


\section{Methodology}
Given a dialogue flow graph $G=(V, E)$ obtained from a dialogue corpus $C=\{C^1, C^2, \dots, C^N \}$, we evaluate the flow from two perspectives: (i) how well does the flow represent the dialogue corpus (information loss) (ii) how well is the representation compressed/denoised (complexity).

\subsection{Information Loss}
We define information loss as the distance between a dialogue corpus and the discovered flow from the corpus, which is the average distance between each dialogue in the corpus and the flow. Intuitively, the more similar the conversations in the corpus are to the paths in the flow, the less the amount of information loss is. More formally, assume that $C^i \in C$ is a dialogue, and $F_G = \{P^k = G_rP_1^kP_2^k...P_{n_k}^k \}_{K}$ is the set of all flow paths in $G$ starting at the root node $G_r$ and ending at a leaf node. The fuzzy dialogue-graph distance between $C^i$ and $G$ can be defined as:
\begin{equation}
\label{eq:fudge}
    FuDGE (C^i, G) = \min_{P^k \in F_G} dist(C^i, P^k) 
\end{equation}

The distance between the dialogue flow G and corpus C is then defined as:
\begin{equation}
\label{eq:avg_fudge}
   f(C, G) = \frac{1}{N} \sum_{i=1}^N FuDGE(C^i, G)
\end{equation}

The distance between a dialogue and a single flow path $dist(C^i, P^k)$ is the edit distance between the two, where each node intent in the flow is paired with an utterance in the conversation through insertion, deletion, or substitution of nodes in the path. Although it follows the same logic as the Levenshtein distance for strings, it poses unique challenges; a dialogue is a sequence of utterances, while a flow path is a sequence of intents (collection of utterances), as opposed to the strings that are sequences of unit characters. The FuDGE algorithm takes these unique characteristics and efficiently calculates the distance between a dialogue and a flow. We describe the detail of this algorithm later in the following sections. 

\subsection{Complexity}
The complexity of the representation can be defined as the complexity of the graph representing the dialogue flow. Depending on the application, there are various ways to calculate the complexity of a graph. Here we define complexity as the number of nodes of a graph. 

\subsection{Flow-F1 Score (\textit{FF1})}
There is a trade-off between the amount of information being represented with the flow and the complexity of the generation. To capture this trade-off, we propose to take the harmonic mean between the normalized complexity and the FuDGE score. To normalize the complexity, we divide it by the total number of utterances, as it is the upper bound for the graph size if we include every single utterance from the dialogue corpus as a node. The maximum FuDGE score is bounded by the average conversation length since the highest score one can get is from an empty graph by inserting every single utterance. Therefore, we normalize the FuDGE score by the average conversation length in the dialogue corpus. The \textbf{F}low-\textbf{F}1 (FF1) is:
\begin{equation}
    FF1 = \frac{2(1-nc)\times(1-nf)}{(1-nc) + (1-nf)}
\end{equation}

where $nc$ and $nf$ are normalized complexity and normalized average FuDGE score (Equation \ref{eq:avg_fudge}). 

\section{Fuzzy Dialogue-Graph Edit Distance}
Motivated from the Levenshtein distance, we focus on aligning a flow path with a dialogue. This is particularly challenging as we need to match a given utterance in the dialogue with an intent in the flow path. Intuitively, an utterance is a good match with an intent if it is semantically close to the majority of the utterances associated with the intent. 

More precisely, given a dialogue flow $G = (V, E)$, its set of flow paths $F_G = \{P^k = G_rP_1^kP_2^k...P_{n_k}^k \}_{K}$, and a conversation $C^i=u^i_1u^i_2\dots u^i_m$, we start with finding the edit distance between $C^i=u^i_1u^i_2\dots u^i_m$ and a specific flow path $P^j = G_rP_1^jP_2^j...P_{n_j}^j$. Conversation $C^i$ is a sequence of utterances $u^i_1u^i_2\dots u^i_m$  produced by a set of actors $a^i_1a^i_2\dots a^i_m$, and flow path $P^j = G_rP_1^jP_2^j...P_{n_j}^j$ is a sequence of nodes, where each node $P_r^j$ is associated with an intent bucket $B^r = (actor, utterances)$. The dialogue flow path edit distance follows the logic described in the preliminaries section. It is similar to Equation \ref{eq:Edit_Distance}, except that we need to define the substitution cost between an utterance and an intent. Once the distance between a single flow path and a dialogue is determined, the FuDGE distance can be computed using the formula in Equation \ref{eq:fudge}. 

\subsection{Fuzzy Substitution Cost}
We want to match an intent with an utterance if the utterance is semantically similar to the 
utterances in the intent bucket.
\begin{equation}
    c_{sub} (B^r, u) \sim d_1 (B^r, u)
\end{equation}

If the utterance and the intent are semantically close to each other, they get matched, but on the other hand, if they are not similar, the intent should be replaced. It is worth noting that we cannot simply replace an intent with an utterance. Therefore, we propose to replace the intent with the nearest intent to the \textit{u} in the set of universal intent buckets $B$. Intuitively, if the current node intent is dissimilar to $u$, it should be replaced with the most similar intent to $u$. Define $B^*$ as the most similar intent to the $u$, therefore: 

\begin{equation}
    cost_{sub} (B^r, u) \sim d_2 (B^r, B^*)
\end{equation}

We define the final substitution cost as:
\begin{equation}
     cost_{sub} (B^r, u) = \alpha (d_1 (B^r, u) + d_2 (B^r, B^*))
\end{equation}

Where $d_1$ and $d_2$ are the intent-utterance and intent-intent distance, respectively, and $\alpha$ is a coefficient set to $0.5$ here to keep the substitution cost between 0 and 1.

We define the intent-utterance and intent-intent distance as a function of their distance in a semantic space. Namely we encode an utterance $u$ and the intent utterances $B^r.utterances = \{u^r_1, \dots u^r_l\}$ into distributional representations $e_u$, and $\{e_1, e_2, \dots, e_l\}$ using Sentence Bert Encoder \cite{reimers2019sentence}. We use the intent centroid as a representation for the intent, obtained by taking the average of the embeddings of all the utterances in the intent, i.e. $e_{B^r} = \frac{1}{l}\sum_{j=1}^{l}e_1$.

Intent-utterance distance is defined in two ways: (i) The cosine distance between an utterance embedding and an intent centroid, and (ii) The cosine distance between the utterance embedding and the nearest utterance in the intent. The intent-intent distance is defined as the cosine distance between the centroids of the two intents. The mathematical formulations are explained in detail in the Appendix.

\subsubsection{Actor Alignment.} An utterance produced by a user should not be matched with an intent associated with an agent and vice versa. Therefore, if actors mismatch, we set the intent-intent and intent-utterance distance to $\infty$. For example: 

\begin{figure}[t]
    \centering
    \includegraphics[width=0.9\columnwidth]{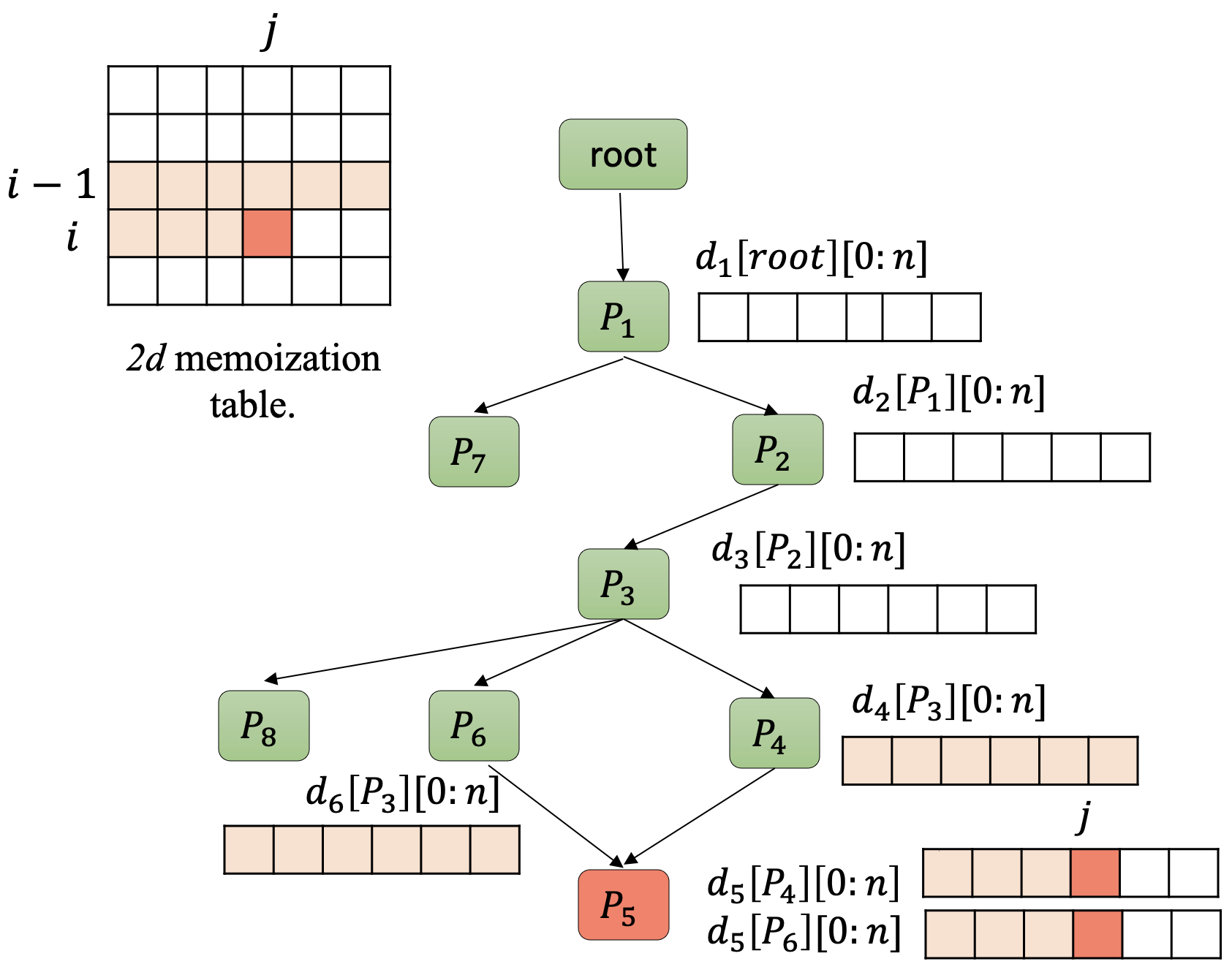}
    \caption{Efficient memoization used for FuDGE}
    \label{fig:memoization}

\end{figure}

\begin{equation}
    d_2 (B^r, B^s)= \\ 
    \left\{
  \begin{array}{@{}ll@{}}
    cosine(e_{B^r}, e_{B^s})  & B^r.actor = B^s.actor \\
    \infty, & \text{otherwise} 
  \end{array}\right.
\end{equation}

\subsection{Complexity Analysis \& Efficient Implementation}
The original edit distance algorithm uses dynamic programming to compute the Levenstein distance. In dynamic programming, the computed solutions to subproblems (e.g., $d_{i, j}$ in Equation \ref{eq:Edit_Distance}) are stored in a memoization table so that these don't have to be recomputed. Even with a memoization table, computing the formula in Equation \ref{eq:fudge} results in a $\mathcal{O}(TKm)$ running time if we separately compute the distance between every single path in the flow $F_G = \{P^k = G_rP_1^kP_2^k...P_{n_k}^k \}_{K}, n_k \leq T$ and a conversation of length $m$. Calculating the distance between a dialogue corpus of size $N$ with a flow graph (Equation \ref{eq:avg_fudge}) results in running time complexity of $\mathcal{O}(TKNm)$. This makes it impractical to calculate the distance between complex dialogue flows with many paths and large dialogue corpora. 

We leverage the structure of a flow DAG to develop a more efficient memoization approach. The core idea behind our approach is that flow paths in a dialogue graph can share overlapping sub-paths. For example in Figure \ref{fig:memoization}, node \texttt{$P_3$} has two children \texttt{$P_4$}, and \texttt{$P_6$}. The two paths \texttt{root->...->$P_3$->$P_4$} and \texttt{root->...->$P_3$->$P_6$} share the same prefix \texttt{root->...->$P_3$}, up to the node  \texttt{$P_3$}, and as a result, both paths require the same memoization information for calculating case 2 and 3 in Equation \ref{eq:Edit_Distance}. Instead of a regular memoization table (top left corner of Figure \ref{fig:memoization}), where row $i$ corresponds to a subsequence ending at index $i$, for each node, we keep an array containing the edit distance between a conversation and a path that starts from the root and ends at the node. Our approach then combines \textit{dfs}-traversal with memoization, where at each node, it uses the parent edit distance array for the calculation in the current node. It is also worth noting that a node might have more than one distance array since multiple paths might end at a certain node (Node $P_5$ in Figure \ref{fig:memoization}). Our approach computes the distance between a dialogue and all the paths in the dialogue flow by taking the entire DAG instead of looking at each path separately. The complexity of computing the FuDGE distance between all flow paths and a conversation with length $n$ is $\mathcal{O}((|V| + |E|)n)$. Algorithm \ref{alg:naive_fudge} and Algorithm \ref{alg:efficient_fudge} provide the implementation detail of naive and efficient algorithms.




\newlength{\origtextfloatsep}
\setlength{\origtextfloatsep}{\textfloatsep}
\setlength{\textfloatsep}{3pt}
\SetKwInput{KwData}{input}
\SetKwInput{KwResult}{output}
\begin{algorithm}[t]
\SetAlgoLined
\definecolor{CommentColor}{HTML}{006619}
\KwData{
    $C^i$ \textcolor{CommentColor}{(Dialogue)};
    $F_G = \{P^k = rP_1^kP_2^k...P_{n_k}^k \}_{k}$ \textcolor{CommentColor}{(A set of dialogue flow paths )};
    \textsc{\texttt{FuzzyEditDistance}} \textcolor{CommentColor}{(function: returns the fuzzy edit distance between a dialogue and a single path) }}
  \SetKwProg{Fn}{def}{:}{}
  \;
  \nl
  \Fn{NaiveFUDGE($C^i$, $F_G$)}
  {
  \nl
  $\texttt{min\_dist} \leftarrow \infty$
  \;
  \nl
  \For{$j \leftarrow 1$ \textbf{to} $k$  }{
  \nl
  $\texttt{dist} \leftarrow \textsc{\texttt{FuzzyEditDistance}}(C^i, P^j)$
  \;
  \nl
  \If{$\texttt{min\_dist} < d$} {
  \nl
    $ \texttt{min\_dist} \leftarrow \texttt{dist}$ 
  \;
}
  }
  \nl
  \textbf{return} $\texttt{min\_dist}$
}
 \caption{Naive Fudge}
\label{alg:naive_fudge}
\end{algorithm}
\SetKwInput{KwData}{input}
\SetKwInput{KwResult}{output}
\begin{algorithm}[t]
\SetAlgoLined
\definecolor{CommentColor}{HTML}{006619}
\KwData{
    $C^i=u_1^iu_2^i...u_n^i$ \textcolor{CommentColor}{(Dialogue)};
    $G = (V, E)$ \textcolor{CommentColor}{(A set of dialogue flow paths )};
    $\texttt{node2dist} \leftarrow \{\}$ \textcolor{CommentColor}{(A map of dialogue flow nodes to a list of tuples (l, d) with l being path length from root to the node and d the path-conversation edit distance.)}; $r \leftarrow G.r$ \textcolor{CommentColor}{(Current node in dfs traversal)}; $p \leftarrow \textsc{\texttt{NaN}}$ \textcolor{CommentColor}{(Current node's parent in dfs traversal)}
    }
    
  \SetKwProg{Fn}{def}{:}{}
  \;
  \nl
  \Fn{FUDGE($C$, $G$, $r$, $p$, $\texttt{node2dist}$)}{
  \nl 
  \textit{DFSEditDistance}$($C$, $G$, $G.r$, \textsc{\texttt{NaN}} , \texttt{node2dist})$
  \;
  \nl
  $\texttt{min\_dist} \leftarrow$ \textit{FindMinPath}$(G, \texttt{node2dist})$
  \;
  \nl
  \textbf{return} $\texttt{min\_dist}$
  }
  \nl
  \;
  \nl
  \Fn{DFSEditDistance($C$, $G$, $r$, $p$, $\texttt{node2dist}$)}
  {
    \If{p = \textsc{\texttt{NaN}}} {
    \nl
    $\texttt{dist} \leftarrow []$
    \;
    \nl
    $n \leftarrow \texttt{len}(C)$
    \;
    \nl
    \For{$i \leftarrow 1$ \textbf{to} $n+1$}{
    \nl
    $\texttt{dist} \leftarrow \texttt{dist} + [i] $
    }
    \nl
    $\texttt{node2dist}[r] \leftarrow [(0, dist)]$
  }
  \nl
  \Else {
  \nl
  \For{l, d \textbf{in} $\texttt{node2dist}[p]$ }{
  \nl
  $\texttt{dist} \leftarrow [l+1]$
  \;
  \nl
  \For{$u \leftarrow u_1^i$ \textbf{to} $u_n^i$}{
  \nl
  $\texttt{dist} \leftarrow \texttt{min}(
  d[i+1] + \texttt{del\_cost}(r.intent),
  \texttt{dist}[-1] + \texttt{insert\_cost(u)},
  d[i] + \texttt{subs\_cost}(r.intent, u))$
  \;
  }
\nl
  $\texttt{node2dist}[r] \leftarrow \texttt{node2dist}[r]+ [(l+1, dist)]$
  }
  \nl
  \For{$c$ \textbf{in} $r.\texttt{children}$}{
  \nl
  \textit{DFSEditDistance}$($C$, $G$, $c$, $r$, \texttt{node2dist})$
  }
  }
  \nl
  \textbf{return}
}

 \caption{Efficient Fudge}
 \label{alg:efficient_fudge}
\end{algorithm}

\section{Experiments}
We describe the automatic flow discovery methods, dialogue corpora, and the evaluation datasets for assessing our study's performance of FuDGE and \textit{FF1}. 
\subsection{Flow Discovery Methods}
Although automatic flow discovery is highly desired, there are only a few discovery algorithms, most unpublished. Graph2Bot \cite{bouraoui-etal-2019-graph2bots} is an algorithm that discovers big convoluted graphs that, although they can be filtered, the over-generation of paths results in loops, which prevent the dialogue flows from being used for a dialogue system. We use two unpublished algorithms (but currently under review) to discover dialogue flows in a dialogue corpus which we call \texttt{ALG1} and \texttt{ALG2}. Both algorithms start with a set of user and agent intents. An intent is a collection of semantically similar utterances representing a user or agent's intention. Both algorithms can use the fully annotated intents when the dialogue corpus is fully annotated. In the absence of human-annotated intents, an intent discovery method is being used to find the intent clusters. \texttt{AGL1} only needs agent intents at the beginning, and the user utterances get grouped as a by-product of graph minimization. For this work, we generate dialogue flows with and without human annotated intents and compare the flows, which we refer to as supervised and unsupervised flows. 

\begin{table*}
\begin{subtable}[t]{0.3\textwidth}
\small
\setlength\tabcolsep{3pt}
    \begin{tabular}[t]{l | c c  }
        \toprule
         & Finance & STAR\\
         \midrule
        Conversations & 447 & 527\\
        Utterances & 7392 & 7352\\
        Tasks & 12 & 5\\
        Agent intents & 55 & 41\\
        User intents & 47 & - \\
        
        \bottomrule
    \end{tabular}
    
\caption{Dataset Statistics}
\label{tab:datasets}
\end{subtable}
\begin{subtable}[t]{0.7\textwidth}
\flushright
\small
\setlength\tabcolsep{2pt}
    \begin{tabular}[t]{l | c c  | c c | c c}
        \toprule
        \multicolumn{1}{c}{} & \multicolumn{2}{c}{Make Payment} & \multicolumn{2}{c}{Hotel Book} & \multicolumn{2}{c}{Bank Report Fraud}\\
        \cmidrule(lr){2-3}
        \cmidrule(lr){4-5}
        \cmidrule(lr){6-7}
        Model & Positves & Negatives & Positves & Negatives & Positves & Negatives  \\
        \midrule
        ALG1-Centroid & $0.14 \pm 0.03$&$0.48 \pm 0.16$&$0.08 \pm 0.03$&$0.59 \pm 0.18$&$0.09 \pm 0.04$&$0.63 \pm 0.19$\\
        ALG1-Min & $0.08 \pm 0.03$&$0.44 \pm 0.16$&$0.04 \pm 0.04$&$0.57 \pm 0.19$&$0.02 \pm 0.04$&$0.60 \pm 0.20$\\
        ALG2-Centroid &$0.27 \pm 0.12$&$0.51 \pm 0.13$&$0.40 \pm 0.21$&$0.63 \pm 0.22$&$0.42 \pm 0.18$&$0.67 \pm 0.22$\\
        ALG1-Min & $0.21 \pm 0.12$&$0.47 \pm 0.13$&$0.34 \pm 0.20$&$0.60 \pm 0.21$&$0.33 \pm 0.18$&$0.61 \pm 0.22$\\
        \bottomrule
    \end{tabular}

\caption{Average FuDGE score for within-task (Positives) vs out-of-task (Negative) conversations, indicating a clear separation.}
\label{tab:e1}
\end{subtable}

\end{table*}

\subsection{Datasets}
 In this work, we use two datasets, each consisting of a set of dialogues, where each dialogue is a conversation between two \textbf{actors} (i.e., user and agent) that consists of a sequence of \textbf{turns}. A turn is an utterance produced by one of the actors. Ideally, the best way to evaluate our framework is to compare a set of manually crafted gold flows, perfect in both coverage and compression, with automatically discovered flows from the dialogue corpus. To our knowledge, there is no public dataset with gold standard flows. Therefore, we propose to impose a level of supervision with human-annotated utterances. Many dialogue state tracking datasets \cite{williams2014dialog, bouraoui-etal-2019-graph2bots, tian2021amendable, qi2022rasat} have human annotated dialogue act utterances, none of which have fully annotated user intents. A dialog act is an utterance that serves as a function in the dialog, such as a question, a statement, or a request.
In contrast, intents are more fine-grained and categorize a user intention. For example, \texttt{"I want to book a hotel room"} and \texttt{"I would like to order pizza"} have the same dialogue act \texttt{request} while their intents are different. Our first dataset, \textbf{Finance}, is a dataset with fully annotated agents and user intents. It constitutes the conversations between a user and the customer service of a financial agency. Our second dataset is created from \textbf{STAR}\cite{mosig2020star}, which is the only publicly available dialogue state tracking dataset partially annotated with agent intents. \textbf{STAR} is a schema-guided task-oriented dialog dataset consisting of 127,833 utterances across 5,820 task-oriented dialogs in 13 domains, from which we pick two domains, \texttt{Hotel} and \texttt{Bank} since they contain the most number of dialogues. We processed the dialogues in the STAR dataset and removed those with unlabeled agent utterances. Table \ref{tab:datasets} contains our datasets' statistics, including the number of dialogues, utterances, tasks, and intents. A complete list of tasks for each dataset is provided in the Appendix.

\subsection{FuDGE Evaluation}
This experiment aims to evaluate the effectiveness of FuDGE as a distance metric. More specifically, given a dialogue flow created for a specific task, a good distance metric should provide lower scores for conversations that belong to the task than the out-of-task conversations. We picked \texttt{Make Payment} task from the Finance dataset with 150 conversations and \texttt{Bank Fraud Report} and \texttt{Hotel Book} from the STAR dataset, with 180 and 145 conversations. We generated separate dialogue flows for each of these tasks using \texttt{ALG1} and \texttt{ALG2}. For each task, we also randomly sampled 50\% of the in-task conversations and added the same number of out-of-task conversations. We evaluated each task-flow with the corresponding dialogue corpus and obtained the average FuDGE score for each dialogue corpus. The results are summarized in Table \ref{tab:e1}. The average score for within-task dialogues (positives) is significantly smaller than the out-of-task (negatives) dialogues. It can segregate within-task conversations from out-of-task conversations. These results suggest that FuDGE is an effective distance metric that can be used independently in any application involving the distance between a dialogue and any predefined DAG structured dialogue scheme. The Appendix provides examples of the dialogues and the best-matched flow path. 

\begin{table*}[t]
\centering
\small
\setlength\tabcolsep{3pt}
    \begin{tabular}{l | c c c c c c | c c c c c c}
        \toprule
        \multicolumn{1}{c}{} & \multicolumn{6}{c}{STAR} & \multicolumn{6}{c}{FINANCE} \\
        \cmidrule(lr){2-7}
        \cmidrule(lr){8-13}
        & \multicolumn{3}{c}{Supervised} & \multicolumn{3}{c|}{Unsupervised} & \multicolumn{3}{c}{Supervised} & \multicolumn{3}{c}{Unsupervised} \\
       \cmidrule(lr){2-4}
       \cmidrule(lr){5-7}
       \cmidrule(lr){8-10}
       \cmidrule(lr){11-13}
        Model & FF1 & FuDGE & Complexity & FF1 & FuDGE & Complexity & FF1 &FuDGE & Complexity & FF1 & FuDGE & Complexity \\
        \midrule
        ALG1-Min  & 0.59 & 0.08 & \multirow{2}{*}{0.57} &\textbf{0.73} & 0.26 & \multirow{2}{*}{0.28} & \textbf{0.71} & 0.03 & \multirow{2}{*}{0.44} & 0.03 & 0.03 & \multirow{2}{*}{0.99}\\
        ALG1-Centroid & 0.58 & 0.12 & &\textbf{0.71} & 0.34 & & \textbf{0.71}& .03  && 0.03 & 0.09 & \\
        \midrule
        ALG2-Min & 0.75 & 0.27 & \multirow{2}{*}{0.23}& \textbf{0.79}& 0.24&\multirow{2}{*}{0.18} & \textbf{0.81} & 0.21 & \multirow{2}{*}{0.18} & 0.67 & 0.23 & \multirow{2}{*}{0.41}\\
        ALG2-Centroid & 0.71 & 0.35 && \textbf{0.73} & 0.34&& \textbf{0.77} & 0.27 && 0.65 & 0.27\\
        \bottomrule
    \end{tabular}
    
\caption{Results of FF1 flow comparison between supervised and unsupervised discovered flow.}
\label{tab:e2}
\vspace{-3pt}
\end{table*}

\subsection{Parameter Optimization With \textit{FF1}}
\begin{figure}[t]
    \centering
    \includegraphics[width=0.85\columnwidth]{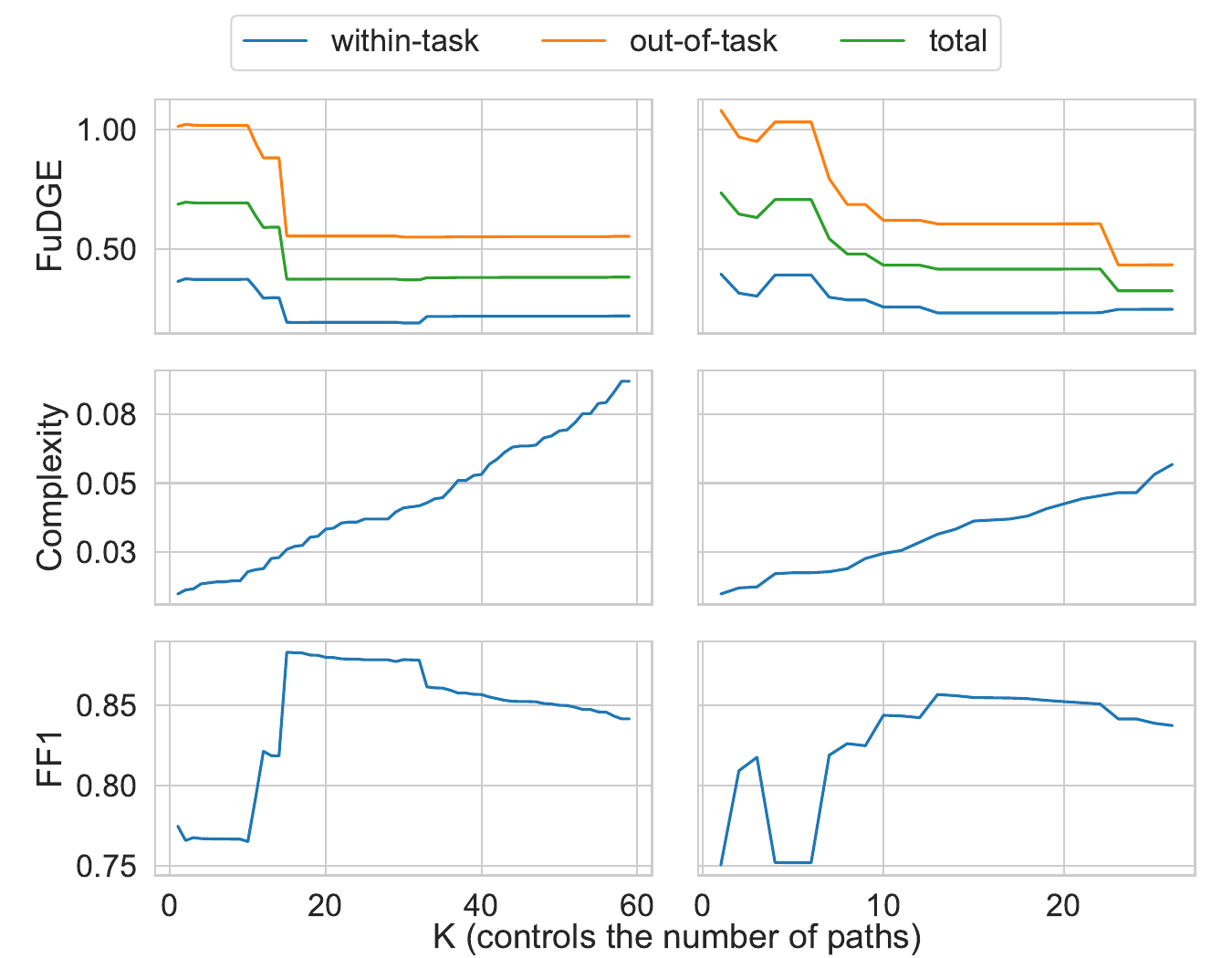}
    \caption{Parameter tuning with \textit{FF1} for \texttt{ALG2} and \texttt{Make Payment} task. The left column is the scores from an unsupervised discovered flow, and the right column corresponds to the supervised flow. The optimal $k$ is smaller for the supervised flow, indicating a better compression.}
\label{fig:e3}
\end{figure}

Automatic flow discovery usually involves multiple steps, including clustering the utterances, creating the graph, ranking important paths, and pruning the graph accordingly. Each of these steps may add different hyperparameters to the entire discovery pipeline. The simplest clustering algorithms, such as K-means, require k as the number of clusters. While hyperparameter selection can drastically impact the quality of the final discovered flows, it has been done mainly by manual trial and error. In this experiment, we show that the \textit{FF1} score is a practical framework for choosing the optimal set of hyperparameters. 
For this experiment, we use \texttt{ALG2} as the flow discovery algorithm. This algorithm consists of a ranking strategy that ranks the paths based on their importance and later keeps the $k$ top-ranked paths as the discovered flow. In both supervised and unsupervised setup, we run \texttt{ALG2} task over the \texttt{Make Payment} task from the Finance dataset. We generate multiple flows for different values of $k$ ranging from 1 to the maximum number of paths in the complete graph. The top row of Figure \ref{fig:e3} shows the normalized FuDGE score obtained for each flow versus $k$. First, we can see the clear segregation between the within-task and out-of-task dialogues. Moreover, as more paths get added, the FuDGE score asymptotically decreases to a minima. The middle row depicts the normalized complexity score for different $k$ values, which indicates a monotonic increase in the complexity, which is expected as we add more paths, and, thus, more nodes to the graph. The bottom row is the final \textit{FF1} score. We see that the scores go up to an optimal point as we add more paths, but it starts declining. The peak in the graph is almost aligned with the point where the FuDGE score starts to stay constant.

\subsection{\textit{FF1} Evaluation}
In this section, we compare dialogue flows with and without annotated intents. The discovery algorithms need to use a clustering method to generate the intents without the labeled data. This process imposes some noise on the flow discovery, leading to lower quality dialogue flows. This experiment aims to show that our evaluation approach can capture this phenomenon. We generated complete flows by running \texttt{ALG1} and \texttt{ALG2} over the entire dialogue corpus for both datasets. Then, we calculated the average FuDGE score and the complexity of each discovered flow and computed the \textit{FF1} score. Table \ref{tab:e2} summarizes the results of this experiment. As shown, for the Finance dataset, the score for the supervised discovered flows is higher than the unsupervised discovered flows, with only one exception. However, the \textit{FF1} score for the unsupervised flows discovered by \texttt{ALG1} is significantly higher than the supervised flows. Manual investigation of the flows showed that the annotated labels were too fine-grained. Clustering led to more high-level intents, which eventually processed better quality dialogue flows. We provide more discussion about this case in the Appendix. 

\section{Related Work}
Related work for our work is relatively sparse. Although automatic evaluation of dialogue systems is an active field of research \cite{yeh2021comprehensive,khalid2022explaining}, most of the metrics and approaches focus on evaluating a dialogue in utterance level \cite{sun2021simulating, ghazarian2020predictive}. However, our work focuses on the evaluation of the dialogues in conversation level, mostly produced by an AI algorithms, such as Graph2Bot introduced by \citet{bouraoui-etal-2019-graph2bots} and is a tool for assisting conversational agent designers. It could extract a graph representation from human-human conversations using unsupervised learning. More recently, \cite{qiu2020structured} used a Variational Recurrent Neural Network model with discrete latent states to learn dialogue structure in an unsupervised fashion. They evaluate their method by using Structure Euclidean Distance (SED) and Structure Cross-Entropy (SCE) based on the transition probabilities between nodes but found them to be unstable. SED and SCE also do not consider the semantic similarity between the node and the original conversation. 

Word Confusion Networks (WCNs) \cite{mangu2000finding} has been used extensively to model the hypothesis of automatic speech recognition (ASR). Just like the dialogue flows, WCNs can also be represented as DAGs. A popular metric for identifying the quality of ASR has been word error rate which incorporates ideas of edit distance that can be derived through each path in the WCN that represents an ASR hypothesis. \citet{lavi2021we} introduced the notion of using edit distance \cite{wagner1974string} for dialog-dialog similarity. In their work, they used sentence-level embeddings \cite{cer2018universal, reimers2019sentence} to determine the similarity between two utterances within a dialogue and defining the edit distance substitution cost.

\section{Conclusion}

This paper presents a novel evaluation framework for a crucial task necessary for building task-oriented dialogue agents. This framework can be used with any flow discovery method and dialogue corpora as long as the generated dialogue flows can be represented as a DAG. We introduced the \textit{FF1} metric, a harmonic mean of flow complexity and FuDGE distance, and demonstrated its efficacy as a tool to select hyperparameters of a flow discovery algorithm or process. We envisage it to be a useful guide for human conversational designers or as a measure to optimize an automatic flow discovery process. We also propose an efficient implementation of FuDGE distance with $\mathcal{O}((|V| + |E|)n)$, allowing it to scale to large datasets. This approach delivers a consistent baseline, thereby better versioning and tracking the progress of flows and corresponding automation with time. In the future, we hope to incorporate utterance characteristics for the insertion and deletion cost to account for the actual semantic cost of the operation.

\bibliography{references}

\begin{thebibliography}{24}
\providecommand{\natexlab}[1]{#1}

\bibitem[{Bouraoui et~al.(2019)Bouraoui, Le~Meitour, Carbou, Rojas~Barahona,
  and Lemaire}]{bouraoui-etal-2019-graph2bots}
Bouraoui, J.-L.; Le~Meitour, S.; Carbou, R.; Rojas~Barahona, L.~M.; and
  Lemaire, V. 2019.
\newblock {G}raph2{B}ots, Unsupervised Assistance for Designing Chatbots.
\newblock In \emph{Proceedings of the 20th Annual SIGdial Meeting on Discourse
  and Dialogue}, 114--117. Stockholm, Sweden: Association for Computational
  Linguistics.

\bibitem[{Cer et~al.(2018)Cer, Yang, Kong, Hua, Limtiaco, John, Constant,
  Guajardo-Cespedes, Yuan, Tar et~al.}]{cer2018universal}
Cer, D.; Yang, Y.; Kong, S.-y.; Hua, N.; Limtiaco, N.; John, R.~S.; Constant,
  N.; Guajardo-Cespedes, M.; Yuan, S.; Tar, C.; et~al. 2018.
\newblock Universal sentence encoder.
\newblock \emph{arXiv preprint arXiv:1803.11175}.

\bibitem[{Devlin et~al.(2018)Devlin, Chang, Lee, and
  Toutanova}]{devlin2018bert}
Devlin, J.; Chang, M.-W.; Lee, K.; and Toutanova, K. 2018.
\newblock Bert: Pre-training of deep bidirectional transformers for language
  understanding.
\newblock \emph{arXiv preprint arXiv:1810.04805}.

\bibitem[{Ester et~al.(1996)Ester, Kriegel, Sander, Xu
  et~al.}]{ester1996density}
Ester, M.; Kriegel, H.-P.; Sander, J.; Xu, X.; et~al. 1996.
\newblock A density-based algorithm for discovering clusters in large spatial
  databases with noise.
\newblock In \emph{kdd}, volume~96, 226--231.

\bibitem[{Forman, Nachlieli, and Keshet(2015)}]{forman2015clustering}
Forman, G.; Nachlieli, H.; and Keshet, R. 2015.
\newblock Clustering by intent: a semi-supervised method to discover relevant
  clusters incrementally.
\newblock In \emph{Joint European Conference on Machine Learning and Knowledge
  Discovery in Databases}, 20--36. Springer.

\bibitem[{Ghazarian et~al.(2020)Ghazarian, Weischedel, Galstyan, and
  Peng}]{ghazarian2020predictive}
Ghazarian, S.; Weischedel, R.; Galstyan, A.; and Peng, N. 2020.
\newblock Predictive engagement: An efficient metric for automatic evaluation
  of open-domain dialogue systems.
\newblock In \emph{Proceedings of the AAAI Conference on Artificial
  Intelligence}, volume~34, 7789--7796.

\bibitem[{Khalid and Lee(2022)}]{khalid2022explaining}
Khalid, B.; and Lee, S. 2022.
\newblock Explaining Dialogue Evaluation Metrics using Adversarial Behavioral
  Analysis.
\newblock In \emph{Proceedings of the 2022 Conference of the North American
  Chapter of the Association for Computational Linguistics: Human Language
  Technologies}, 5871--5883.

\bibitem[{Lavi et~al.(2021)Lavi, Rabinovich, Shlomov, Boaz, Ronen, and
  Anaby-Tavor}]{lavi2021we}
Lavi, O.; Rabinovich, E.; Shlomov, S.; Boaz, D.; Ronen, I.; and Anaby-Tavor, A.
  2021.
\newblock We've had this conversation before: A Novel Approach to Measuring
  Dialog Similarity.
\newblock \emph{arXiv preprint arXiv:2110.05780}.

\bibitem[{Levenshtein et~al.(1966)}]{levenshtein1966binary}
Levenshtein, V.~I.; et~al. 1966.
\newblock Binary codes capable of correcting deletions, insertions, and
  reversals.
\newblock In \emph{Soviet physics doklady}, volume~10, 707--710. Soviet Union.

\bibitem[{Lin, Xu, and Zhang(2020)}]{lin2020discovering}
Lin, T.-E.; Xu, H.; and Zhang, H. 2020.
\newblock Discovering new intents via constrained deep adaptive clustering with
  cluster refinement.
\newblock In \emph{Proceedings of the AAAI Conference on Artificial
  Intelligence}, volume~34, 8360--8367.

\bibitem[{Mangu, Brill, and Stolcke(2000)}]{mangu2000finding}
Mangu, L.; Brill, E.; and Stolcke, A. 2000.
\newblock Finding consensus in speech recognition: word error minimization and
  other applications of confusion networks.
\newblock \emph{Computer Speech \& Language}, 14(4): 373--400.

\bibitem[{Mosig, Mehri, and Kober(2020)}]{mosig2020star}
Mosig, J.~E.; Mehri, S.; and Kober, T. 2020.
\newblock Star: A schema-guided dialog dataset for transfer learning.
\newblock \emph{arXiv preprint arXiv:2010.11853}.

\bibitem[{Perkins and Yang(2019)}]{perkins-yang-2019-dialog}
Perkins, H.; and Yang, Y. 2019.
\newblock Dialog Intent Induction with Deep Multi-View Clustering.
\newblock In \emph{Proceedings of the 2019 Conference on Empirical Methods in
  Natural Language Processing and the 9th International Joint Conference on
  Natural Language Processing (EMNLP-IJCNLP)}, 4016--4025. Hong Kong, China:
  Association for Computational Linguistics.

\bibitem[{Qi et~al.(2022)Qi, Tang, He, Wan, Zhou, Wang, Zhang, and
  Lin}]{qi2022rasat}
Qi, J.; Tang, J.; He, Z.; Wan, X.; Zhou, C.; Wang, X.; Zhang, Q.; and Lin, Z.
  2022.
\newblock RASAT: Integrating Relational Structures into Pretrained Seq2Seq
  Model for Text-to-SQL.
\newblock \emph{arXiv preprint arXiv:2205.06983}.

\bibitem[{Qiu et~al.(2020)Qiu, Zhao, Shi, Liang, Shi, Yuan, Yu, and
  Zhu}]{qiu2020structured}
Qiu, L.; Zhao, Y.; Shi, W.; Liang, Y.; Shi, F.; Yuan, T.; Yu, Z.; and Zhu,
  S.-C. 2020.
\newblock Structured attention for unsupervised dialogue structure induction.
\newblock \emph{arXiv preprint arXiv:2009.08552}.

\bibitem[{Reimers and Gurevych(2019)}]{reimers2019sentence}
Reimers, N.; and Gurevych, I. 2019.
\newblock Sentence-bert: Sentence embeddings using siamese bert-networks.
\newblock \emph{arXiv preprint arXiv:1908.10084}.

\bibitem[{Shi et~al.(2018)Shi, Chen, Sha, Li, Sun, Wang, and
  Zhang}]{shi-etal-2018-auto}
Shi, C.; Chen, Q.; Sha, L.; Li, S.; Sun, X.; Wang, H.; and Zhang, L. 2018.
\newblock Auto-Dialabel: Labeling Dialogue Data with Unsupervised Learning.
\newblock In \emph{Proceedings of the 2018 Conference on Empirical Methods in
  Natural Language Processing}, 684--689. Brussels, Belgium: Association for
  Computational Linguistics.

\bibitem[{Sun et~al.(2021)Sun, Zhang, Balog, Ren, Ren, Chen, and
  de~Rijke}]{sun2021simulating}
Sun, W.; Zhang, S.; Balog, K.; Ren, Z.; Ren, P.; Chen, Z.; and de~Rijke, M.
  2021.
\newblock Simulating user satisfaction for the evaluation of task-oriented
  dialogue systems.
\newblock In \emph{Proceedings of the 44th International ACM SIGIR Conference
  on Research and Development in Information Retrieval}, 2499--2506.

\bibitem[{Tian et~al.(2021)Tian, Huang, Lin, Bao, He, Yang, Wu, Wang, and
  Sun}]{tian2021amendable}
Tian, X.; Huang, L.; Lin, Y.; Bao, S.; He, H.; Yang, Y.; Wu, H.; Wang, F.; and
  Sun, S. 2021.
\newblock Amendable generation for dialogue state tracking.
\newblock \emph{arXiv preprint arXiv:2110.15659}.

\bibitem[{Wagner and Fischer(1974)}]{wagner1974string}
Wagner, R.~A.; and Fischer, M.~J. 1974.
\newblock The string-to-string correction problem.
\newblock \emph{Journal of the ACM (JACM)}, 21(1): 168--173.

\bibitem[{Williams et~al.(2014)Williams, Henderson, Raux, Thomson, Black, and
  Ramachandran}]{williams2014dialog}
Williams, J.~D.; Henderson, M.; Raux, A.; Thomson, B.; Black, A.; and
  Ramachandran, D. 2014.
\newblock The dialog state tracking challenge series.
\newblock \emph{AI Magazine}, 35(4): 121--124.

\bibitem[{Yeh, Eskenazi, and Mehri(2021)}]{yeh2021comprehensive}
Yeh, Y.-T.; Eskenazi, M.; and Mehri, S. 2021.
\newblock A comprehensive assessment of dialog evaluation metrics.
\newblock \emph{arXiv preprint arXiv:2106.03706}.

\bibitem[{Zhang et~al.(2021)Zhang, Xu, Lin, and Lyu}]{zhang2021discovering}
Zhang, H.; Xu, H.; Lin, T.-E.; and Lyu, R. 2021.
\newblock Discovering new intents with deep aligned clustering.
\newblock In \emph{Proceedings of the AAAI Conference on Artificial
  Intelligence}, volume~35, 14365--14373.

\bibitem[{Zhang et~al.(2022)Zhang, Zhang, Zhan, Wu, and Lam}]{zhang2022new}
Zhang, Y.; Zhang, H.; Zhan, L.-M.; Wu, X.-M.; and Lam, A. 2022.
\newblock New Intent Discovery with Pre-training and Contrastive Learning.
\newblock \emph{arXiv preprint arXiv:2205.12914}.

\end{thebibliography}

\appendix
\clearpage
\section{Mathematical Formulations}
This section provides the exact mathematical formulation to calculate the fuzzy substitution cost. For the representation of an intent bucket, we average the embedding of the utterances in the bucket. 
\begin{equation}
   e_{B^r} = \frac{1}{|B^r.utterances|} \sum_{u^j \in B^r.utterances}e_{u^j} 
   \label{eq:intent_centroid}
\end{equation}

The distance between an intent $B^r$ and utterance $u$ is $d_1$ and is defined as: 
\begin{equation}
    d_1 (B^r, u)= \\ 
    \left\{
  \begin{array}{@{}ll@{}}
    \phi(B^r, u)  & B^r.actor = u.actor \\
    \infty, & \text{otherwise} 
  \end{array}\right.
\end{equation}

Where $\phi$ is the function of intent and utterance semantic embeddings and is defined in two following ways:

\begin{enumerate}
    \item \textbf{Min}. The minimum distance between $u$ and the nearest utterance in $B^r.utterances$. 
    \[\phi (B^r, u) = \min_{u^j \in B^r.utterances} cosine(e_{u^j}, e_{u})\]
    \item \textbf{Centroid}. The cosine distance between $e_{B^r}$ and $e_u$
    \[ \phi (B^r, u) = cosine(e_u, e_{B^r}) \]
\end{enumerate}

For the intent-intent distance $d_2$ we use the formula in Equation \ref{eq:intent_centroid} to obtain two intent representation, $d_2$ then is:

\begin{equation}
    d_2 (B^r, B^s)= \\ 
    \left\{
  \begin{array}{@{}ll@{}}
    cosine(e_{B^r}, e_{B^s})  & B^r.actor = B^s.actor \\
    \infty, & \text{otherwise} 
  \end{array}\right.
\end{equation}

While the \textbf{Min} approach often produces smaller FuDGE score, and therefor larger \textit{FF1} (Table \ref{tab:e1}, Table \ref{tab:e2}), the \textbf{Centroid} method is more robust to the effect of outliers in the intent clusters. 

\section{Dataset Detail}
Table \ref{tab:data_topics} contains the dataset tasks and the number of conversations per task. The tasks with the most number of conversations are selected for the second and third sections in the experiments section. 
\begin{table}[t]
\centering
    \begin{tabular}{l c}
        \toprule
        \multicolumn{2}{c}{Finance}\\
        \midrule
        \texttt{Make Payment} & $150$\\
        \texttt{Investigate Charges} & $66$\\
        \texttt{Check Card Balance} & $51$\\
        \texttt{Credit Limit Request} & $36$\\
        \texttt{Replace Card} & $36$\\
        \texttt{Loan Status Request} & $24$\\
        \texttt{Lock Card Request} & $24$\\
        \texttt{Lost Card } & $12$\\
        \texttt{Activate Card} & $12$\\
        \texttt{Lost Card} & $12$\\
        \texttt{Unlock Card} & $12$\\
        \texttt{Card Status} & $12$\\
        \midrule
        \multicolumn{2}{c}{STAR}\\
        \midrule
        \texttt{Bank Fraud Report} & $180$\\
        \texttt{Hotel Book} & $145$\\
        \texttt{Hotel Search} & $108$\\
        \texttt{Bank Balance} & $43$\\
        \texttt{Hotel Service Request} & $31$\\
        \bottomrule
    \end{tabular}
    
\caption{Tasks in each datasets with the number of conversation within each task}
\label{tab:data_topics}

\end{table}

We are publishing the STAR dataset, including the dialogue corpora used for flow discovery and evaluation, and the discovered flows from the STAR dialogue corpus using \texttt{ALG1} and \texttt{ALG2}. 

\section{Flow Investigation}
In this section, we provide a more detailed discussion of the results in Table \ref{tab:e2} and the reason why for the STAR dataset, the ff1 score is higher for the unsupervised flows compared to the supervised flows. After manual investigation, we concluded that the agent labels were too fine-grained. Table \ref{tab:cluster_intents} is providing an example of this scenario. The three intents in the supervised setup are clustered together by the clustering algorithm in the unsupervised setup. This is often the case that these three intents occur in a different order within a conversation; therefore, having one intent instead of three results in a more compact denoised flow.

In the supplementary material in folder \texttt{nfa\_flows}, we are providing a visualization of the flows, generated by \texttt{ALG2}. We provide flow visualization for both supervised and unsupervised settings referred to as \texttt{labeled} and \texttt{unlabeled}. 
\begin{table*}[t]
\small
\setlength\tabcolsep{3pt}
\centering
    \begin{tabular}{l l l }
        \toprule
        Unsupervised Intent & Supervised Intent & Utterances \\
        \midrule
        \multirow{3}{*}{\texttt{date birth}} & \texttt{bank\_ask\_mothers\_maiden\_name} & \texttt{What was your mother's maiden name?}\\
         & \texttt{bank\_ask\_dob} & \texttt{Could you provide your date of birth, please?}\\
        &\texttt{bank\_ask\_childhood\_pet\_name} & \texttt{And what was the name of the pet you had as a child?}\\
        \bottomrule
    \end{tabular}
    
\caption{Tasks in each datasets with the number of conversation within each task}
\label{tab:cluster_intents}
\vspace{-3pt}
\end{table*}

\section{Examples Of Matched Paths}
\label{sec:examples}
This section provides some examples of conversations, the path picked up by the FuDGE algorithm, and the operations and costs needed to convert a path to a conversation. The best-matched path to a conversation is a path selected from the paths that start from the root node, end at one of the leaf nodes, and have the lowest FuDGE score. To find the best-matched paths, we look into the distance array kept for each leaf node at \texttt{node2dist} map. (refer to Algorithm \ref{alg:efficient_fudge}, and we peak the node with the smallest edit distance at \texttt{node2dist[leaf][-1]}. After we picked the leaf node with the smallest distance, we tracked the path by reversing the steps from the child node to the parent node until the root node. Table \ref{tab:alignment} contains examples of three conversations with their best-matched paths. More examples can be found in the supplementary directory \texttt{alignment}.

\begin{table*}[t]
\small
\setlength\tabcolsep{3pt}
\centering
    \begin{tabular}{V{10cm} V{4cm} l l }
        \toprule
        Conversation & Path Intent Names & Operations & Cost \\
        \midrule
        \textbf{u.}\texttt{Hi there, I need to reserve a hotel room!}&\texttt{Reserve Hotel Room}&\texttt{replace}&0.346\\
\textbf{a.}\texttt{What hotel would you like to stay at?}&\texttt{Hotel Like Stay}&\texttt{replace}&0.346\\
\textbf{u.}\texttt{Good question.  I wanted to say the Hilton, but my friend recommends the Old Town Inn, so lets try that}&\texttt{Town Inn}&\texttt{replace}&0.407\\
\textbf{a.}\texttt{When are you arriving?}&\texttt{Arriving Arriving}&\texttt{replace}&0.407\\
\textbf{u.}\texttt{12-May}&\texttt{May 12 Arrive}&\texttt{replace}&0.513\\
\textbf{a.}\texttt{When will you be leaving again?}&\texttt{Leaving Leaving}&\texttt{replace}&0.513\\
\textbf{u.}\texttt{Actually never mind the Old Town Inn, my personal favorite blog says the Hyatt is the bees knees.  Let's do that instead}&\texttt{Hyatt Bees Knees}&\texttt{replace}&0.513\\
\textbf{a.}\texttt{When will you be leaving again?}&\texttt{When Will You}&\texttt{replace}&0.513\\
\textbf{u.}\texttt{Oh yeah the 24th, this blog is the bomb!}&\texttt{24Th Blog Bomb}&\texttt{replace}&0.513\\
\textbf{a.}\texttt{May I have your name, please?}&\texttt{May Name Please}&\texttt{replace}&0.540\\
\textbf{u.}\texttt{Would you believe this.... my wife just sent me a text saying my brother in law is getting married in London, ironically on the 24th... so scratch this month and lets do the 8th to the 26th next month.}&\texttt{Getting Married London}&\texttt{replace}&0.540\\
\textbf{a.}\texttt{May I have your name, please?}&\texttt{May Name Please}&\texttt{replace}&0.566\\
\textbf{u.}\texttt{Oh yeah sorry Ben with a B}&\texttt{Yeah Sorry Ben}&\texttt{replace}&0.566\\
\textbf{a.}\texttt{Alright, the Hyatt Hotel ticks all of your boxes, can I book this room for you?}&\texttt{Alright Hyatt Hotel}&\texttt{replace}&0.566\\
\textbf{u.}\texttt{Yes please.  Let's be honest here nobody really likes weddings right?}&\texttt{Really Likes Weddings}&\texttt{replace}&0.566\\
\textbf{a.}\texttt{OK, I've successfully completed this Hotel booking for you!}&\texttt{Successfully Completed Hotel}&\texttt{replace}&0.566\\
\textbf{u.}\texttt{Ok great thanks a lot}&\texttt{Ok Great Thanks}&\texttt{replace}&0.886\\
\midrule
\midrule
\textbf{a.}\texttt{Hello I need to reserve a room. My friend is having a big party.}&\texttt{town inn}&\texttt{replace}&0.383\\
\textbf{u.}\texttt{Hello}&\texttt{Hello Hello Hello}&\texttt{replace}&0.387\\
\textbf{a.}\texttt{Hello, how can I help?}&\texttt{Hello Help}&\texttt{replace}&0.387\\
\textbf{u.}\texttt{I need to reserve a room. My friend is having a big party.}&\texttt{Want Resevation}&\texttt{replace}&0.959\\
\textbf{a.}\texttt{May I have your name, please?}&\texttt{May Name Please}&\texttt{replace}&0.985\\
\textbf{u.}\texttt{Angela}&\texttt{John Angela Alexis}&\texttt{replace}&1.259\\
\textbf{a.}\texttt{What hotel would you like to stay at?}&\texttt{Hotel Like Stay}&\texttt{replace}&1.259\\
\textbf{u.}\texttt{Old Town Inn is my favorite. Hopefully it is available.}&\texttt{Hilton Hyatt Hyatt}&\texttt{replace}&1.826\\
\textbf{a.}\texttt{When are you arriving?}&\texttt{Arriving Arriving Arriving}&\texttt{replace}&1.826\\
\textbf{u.}\texttt{May 8th. It is also my birthday. I am a stubborn Taurus.}&\texttt{Arriving 11Th}&\texttt{replace}&2.319\\
\textbf{a.}\texttt{When will you be leaving again?}&\texttt{When Will You}&\texttt{replace}&2.319\\
\textbf{u.}\texttt{May 23rd I will be leaving.}&\texttt{Request Extra Towels}&\texttt{replace}&2.759\\
\textbf{a.}\texttt{Do you have any special requests?}&\textcolor{blue}{\texttt{Do you have any special requests?}}&\texttt{insert}&3.759\\
\textbf{u.}\texttt{No. I am a simple earth sign.}&\textcolor{blue}{\texttt{No. I am a simple earth sign.}}&\texttt{insert}&4.759\\
\textbf{a.}\texttt{I'm very sorry, but there is no room available at the Old Town Inn for your requested dates.}&\texttt{Hotels Match Search}&\texttt{replace}&5.298\\
\textbf{u.}\texttt{That is okay Thanks for trying. Goodbye.}&\texttt{Birth Hospital Goodbye}&\texttt{replace}&5.893\\
\textbf{a.}\texttt{Thank you and goodbye.}&\texttt{Thank Goodbye}&\texttt{replace}&5.893\\
        \bottomrule
    \end{tabular}
    
\caption{Alignment of a conversation with a flow path. Intent names are generated with NGrams from the a cluster utterance set}
\label{tab:alignment}
\vspace{-3pt}
\end{table*}

\end{document}